\documentclass[cameraready]{Interspeech}
\usepackage{graphicx}
\usepackage{booktabs}
\usepackage{multirow}
\usepackage{amsmath, amssymb}
\usepackage{url}
\title{Rubric-Aligned Disentangled Evaluation of Human Simultaneous Interpreting}
\author[affiliation={1},  equalcontribution]{Ziyu}{Zhang}
\author[affiliation={2,3}, equalcontribution, correspondingauthor]{Satoshi}{Nakamura}

\address{
    $^1$ School of Data Science, The Chinese University of Hong Kong, Shenzhen, China \\
    $^2$ School of Artificial Intelligence, The Chinese University of Hong Kong, Shenzhen, China \\
    $^3$ Shenzhen Loop Area Institute, China
}
\email{ziyuzhang@cuhk.edu.cn, snakamura@cuhk.edu.cn}

\begin{document}
\maketitle
\begin{abstract}
Human simultaneous interpreting (SI) is commonly assessed with analytic rubrics separating meaning transfer, delivery quality, and temporal synchrony, yet no automatic metric is designed for rubric-aligned segment-level SI evaluation. We construct a professionally annotated corpus of 1,101 SI segments with scores for meaning transfer (LQ), delivery quality (EXP), and perceived latency (LAT). We show that structured LLM prompting and scalar supervision collapse rubric dimensions, yielding near-zero correlation with human ratings and strong cross-dimension coupling. To isolate supervision structure under identical backbone capacity, we introduce dual regression heads on a LoRA-adapted COMET-KIWI encoder. On a held-out talk-level test set, the model achieves Pearson correlations of 0.388 (LQ) and 0.301 (EXP), improving over frozen COMET-KIWI. Given low absolute rater agreement, we interpret results relative to human consistency and target stable ranking signals for formative assessment.
\end{abstract}

\section{Introduction}

Human simultaneous interpreting (HSI) involves professional interpreters producing target speech in real time while listening to the source. In professional and educational contexts, HSI performance is commonly assessed using analytic rubrics that separately evaluate meaning transfer, delivery quality, and temporal synchrony. In this work, we operationalize these dimensions at the segment level as \textbf{LQ} (semantic fidelity), \textbf{EXP} (delivery quality), and \textbf{LAT} (perceived synchrony).

Professional certification frameworks such as NAATI~\cite{naati2024} and CATTI~\cite{catti2023} employ multi-criteria analytic scoring that distinguishes semantic fidelity from delivery quality, while recent proposals such as SVIP~\cite{cheng2025seedliveinterpret20endtoend} further formalize structured multi-dimensional SI evaluation. However, these frameworks operate at the interpreter level and do not provide fine-grained segment-level diagnostics, and SI assessment remains labor-intensive and subject to inter-rater variability due to subjective judgment~\cite{Fantinuoli2018}, creating a gap between professional standards and scalable automatic evaluation.

Existing MT evaluation metrics—including BLEU, METEOR, BERTScore, BLEURT, COMET, and COMET-KIWI—were developed for written translation and typically collapse quality into a single scalar score~\cite{papineni-etal-2002-bleu, banerjee-lavie-2005-meteor, zhang2020bertscoreevaluatingtextgeneration, sellam-etal-2020-bleurt, rei-etal-2020-comet, rei2022cometkiwiistunbabel2022submission}. However, SI differs fundamentally from written translation due to incremental processing, temporal constraints, and frequent reformulation or omission~\cite{gile2009book, setton2016conference}. Prior work further reports discrepancies between MT-oriented metrics and human SI judgments under interpreting-specific phenomena such as summarization and latency effects~\cite{zhang2021bstclargescalechineseenglishspeech, wein-etal-2024-barriers}. 

Recent work explores large language models (LLMs) as automatic evaluators for text generation~\cite{liu-etal-2023-g, li2023camelcommunicativeagentsmind, li2024llmsasjudgescomprehensivesurveyllmbased}. We evaluate structured zero-shot and few-shot prompting strategies that enforce explicit scoring order (``LQ first, then EXP''). Nevertheless, LLM predictions exhibit strong cross-dimensional coupling (corr $\approx 0.90$ on dev), suggesting that instruction-level control alone fails to preserve rubric structure.

Motivated by this observation, we hypothesize that under correlated analytic dimensions and rater-specific scale variability, scalar supervision collapses ranking structure even with identical encoder capacity. We therefore test whether parameter-level structural separation is necessary to preserve rubric-aligned signals.

\begin{figure}[t]
  \centering
  \includegraphics[width=\linewidth]{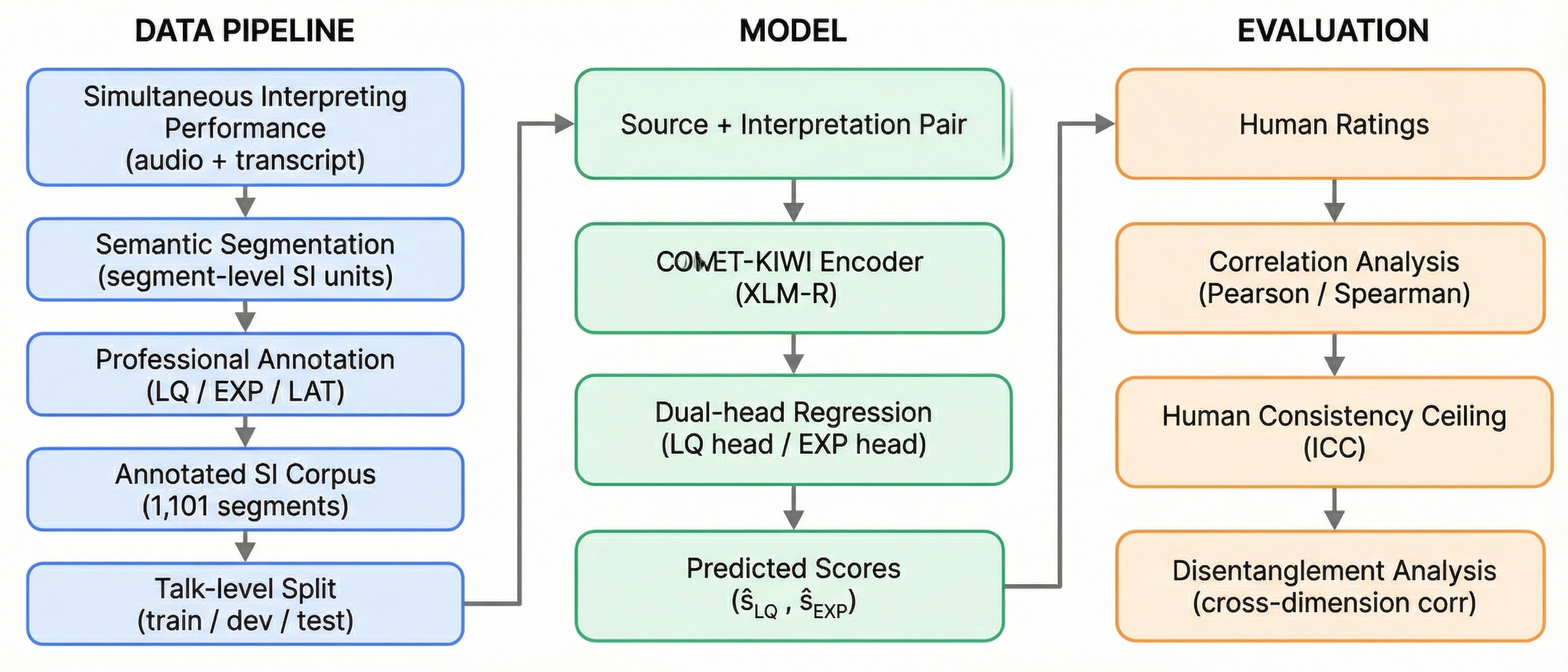}
  \caption{Overview of the rubric-aligned SI evaluation framework: segment-level rubric annotation (LQ/EXP/LAT) with talk-level split, dual-head COMET-KIWI for LQ/EXP, and evaluation via correlation, human reliability, and disentanglement.}
  \label{fig:pipeline}
\end{figure}
To address this, we introduce a rubric-aligned neural evaluation framework that explicitly separates meaning transfer and delivery quality. We adopt COMET-KIWI~\cite{rei2022cometkiwiistunbabel2022submission} as a reference-free quality estimation backbone and adapt it to rubric supervision via lightweight LoRA~\cite{hu2021loralowrankadaptationlarge} and disentangled regression heads (Figure~\ref{fig:pipeline}). Our contributions are threefold:
\begin{enumerate}
\item A professionally annotated segment-level SI corpus (1,101 segments) with analytic rubric supervision.
\item Empirical evidence that prompt-based and scalar supervision collapse rubric dimensions.
\item A dual-head neural metric that improves correlation with professional ratings and yields stable ranking signals under noisy supervision.
\end{enumerate}

Although we use a dual-head architecture, the central challenge is supervision structure under correlated-but-distinct rating dimensions, not conventional multi-task learning. Our goal is not to replace certification assessment, but to support formative segment-level feedback interpreted relative to human consistency.

\section{Related Work}

\subsection{Automatic Metrics for Machine Translation}

Most automatic evaluation metrics were developed for written machine translation and assume scalar quality representation. Classical metrics such as BLEU, TER, METEOR, and chrF ~\cite{papineni-etal-2002-bleu,snover-etal-2006-study,banerjee-lavie-2005-meteor,popovic-2015-chrf} rely on surface similarity, while neural metrics including BERTScore, BLEURT, YiSi, COMET and COMET-KIWI ~\cite{zhang2020bertscoreevaluatingtextgeneration,sellam-etal-2020-bleurt,lo-2019-yisi,rei-etal-2020-comet,rei2022cometkiwiistunbabel2022submission}, leverage pretrained multilingual encoders to approximate human adequacy judgments. 

\subsection{Evaluation of Human Simultaneous Interpreting}

HSI is produced incrementally under real-time constraints and is commonly analyzed through latency--quality trade-offs~\cite{elbayad2020waitk}, often operationalized using measures such as Ear--Voice Span (EVS), a classical latency measure in simultaneous interpreting~\cite{goldman1972segmentation}. Analyses of simultaneous interpretation corpora, including large-scale sentence-aligned datasets~\cite{doi-etal-2021-large}, as well as speech translation resources such as BSTC, report discrepancies between MT-oriented metric scores and human SI judgments under interpreting-specific phenomena such as summarization and reformulation~\cite{wein-etal-2024-barriers}.

Professional assessment frameworks such as NAATI and CATTI employ analytic scoring that separates meaning transfer and delivery quality, while academic proposals such as SVIP~\cite{cheng2025seedliveinterpret20endtoend} further formalize multi-dimensional SI evaluation principles under structured rubric design. However, prior work lacks neural evaluation models explicitly trained under rubric supervision to disentangle these dimensions in human SI.

\subsection{LLM-Based Evaluation of Generated Text}

Recent work explores LLMs as evaluators for text generation and SI, including GPT-3.5-based SI evaluation, prompt-based scoring, and multi-agent debate frameworks~\cite{liu-etal-2023-g, fantinuoli-wang-2024-exploring, li2024llmsasjudgescomprehensivesurveyllmbased}. Unlike reference-based machine-SI evaluation, our setting is reference-free human SI evaluation under professional analytic rubrics. Such LLM-based evaluation relies mainly on instruction-level control, raising concerns about rubric-level controllability.

\section{Dataset and Rubric}
The corpus combines publicly available BSTC-derived SI segments ~\cite{zhang2021bstclargescalechineseenglishspeech} with newly collected licensed TED-style conference recordings obtained under research consent. Transcripts were obtained from official transcripts where available and from ASR for collected recordings, followed by normalization. After combining all sources, we performed the train/dev/test split at the talk level to prevent information leakage across related discourse~\cite{roberts2017crossvalidation}.

\subsection{Dataset and Split}
\begin{table}[t]
\caption{Segment-level rubric summary (0--3). }
\label{tab:rubric}
\centering
\small
\begin{tabular}{lp{0.78\linewidth}}
\toprule
Dim. & Criteria (summary) \\
\midrule
LQ & 3: $>$80\% meaning preserved; no major omission/distortion. 
2: $\sim$60--70\% meaning; some omissions/inaccuracies but main message intact.
1: $\sim$40--50\% meaning; major omissions/distortions. 
0: $<$30\% meaning or misleading/reversed meaning. \\
EXP & 3: fluent, idiomatic; minor errors only. 
2: generally fluent with noticeable awkwardness/errors.
1: frequent disfluency/grammar issues affecting delivery.
0: breakdown-level disfluency impedes comprehension. \\
LAT & 3: low perceived lag; well synchronized.
2: occasional lag; manageable.
1: frequent noticeable lag; disjointed flow.
0: severe persistent lag; difficult to follow. \\
\bottomrule
\end{tabular}
\end{table}

Each segment is double-rated by professional raters under the analytic rubric (Table\ref{tab:rubric}), and individual ratings are treated as separate supervision instances. The resulting split comprises 839 segments (48 talks) for training, 87 segments (5 talks) for development, and 169 segments (8 talks) for testing, covering both English$\rightarrow$Chinese and Chinese$\rightarrow$English directions. Pre-processing included semantic segmentation of source–interpretation pairs, transcript normalization, and removal of incomplete or unavailable segments prior to splitting.

\begin{table}[t]
\caption{Dataset statistics and talk-level split.}
\label{tab:data_stats}
\centering
\small
\begin{tabular}{lccc}
\toprule
 & Train & Dev & Test \\
\midrule
\#Segments & 839 & 87 & 169 \\
\#Talks    & 48  & 5  & 8   \\
\midrule
Split level & \multicolumn{3}{c}{Talk (no cross-talk overlap)} \\
Directions  & \multicolumn{3}{c}{En$\rightarrow$Zh and Zh$\rightarrow$En} \\
\bottomrule
\end{tabular}
\end{table}

\subsection{From Professional Framework to Segment-Level Operationalization}

Professional SI certification frameworks (e.g., NAATI, CATTI) and analytic proposals such as SVIP evaluate interpreting performance along separable dimensions including meaning transfer and delivery quality. Building on these rubric-based frameworks, we introduce a modeling-oriented segment-level operationalization that preserves analytic separation while enabling neural supervision. Scoring is performed on semantically segmented units, a deterministic gating rule sets EXP to zero when LQ indicates severe meaning failure, and latency is annotated impressionistically to capture perceived synchrony rather than precise acoustic timing.

\subsection{Latency Dimension Analysis}
\label{sec:latency_analysis}

The annotation protocol includes a perceived latency (LAT) dimension scored on a 0--3 scale alongside objective segment-level onset delay. Perceived latency shows substantial variation (mean = 2.13, std = 0.79) but weak correlation with objective delay (Pearson = -0.048; Spearman = -0.034), suggesting discourse-level synchrony judgments consistent with cognitive-load–based accounts of simultaneous interpreting~\cite{gile2009effort, kano2023averagetokendelaylatency}. LAT additionally correlates with LQ ($r=0.38$) and EXP ($r=0.43$), indicating a latency--quality trade-off.

\section{Method}

We model rubric-aligned SI evaluation as multi-output regression,
$f(x,y)\rightarrow(\hat{s}_{LQ},\hat{s}_{EXP})$,
where $x$ denotes the source and $y$ the interpreted output.
The COMET-KIWI encoder was initialized from pretrained weights,
while regression heads were randomly initialized using scaled Xavier initialization.
We focus on text-only prediction of LQ and EXP. Since EXP also depends on acoustic cues such as prosody, pauses, and fluency, our model is a text-based lower bound that captures transcript-visible delivery signals. Modeling LAT likely requires multimodal timing cues and is left for future work.

Our approach relies on three assumptions:
(i) rubric dimensions such as meaning transfer and delivery quality are analytically separable despite empirical correlation;
(ii) individual rater supervision better captures scale variability than aggregated mean labels; and
(iii) formative SI assessment primarily requires stable ranking signals rather than absolute agreement with any single rater.

\subsection{Backbone and Dual-Head Architecture}

We build on COMET-KIWI~\cite{rei2022cometkiwiistunbabel2022submission} (XLM-R Large) using pair encoding
\texttt{[CLS] source [SEP] hypothesis [SEP]}.
The original scalar regression head is replaced with two independent linear heads:
$\hat{s}_{LQ}=W_{LQ}h+b_{LQ}$ and
$\hat{s}_{EXP}=W_{EXP}h+b_{EXP}$,
where $h$ denotes the [CLS] representation.
The backbone contains approximately 550M parameters, while LoRA and regression heads introduce fewer than 1M additional trainable parameters.

\subsection{Residual Prediction and Objective}

To mitigate prediction mean-collapse under noisy supervision, we adopt residual prediction
$\hat{s}_{d}=\mu_d+\Delta_d$ for $d\in\{LQ,EXP\}$ and optimize

\[
\mathcal{L}=\mathrm{MSE}_{LQ}+w\,\mathrm{MSE}_{EXP}+\lambda\mathcal{L}_{var},
\]

with $w{=}1.7$ and $\lambda{=}0.05$.
Predictions are clamped to $[0,3]$ for evaluation (optionally quantized to 0.5 steps).

\subsection{LoRA Adaptation}

Encoder adaptation is performed using LoRA~\cite{hu2021loralowrankadaptationlarge} applied to attention $Q,V$ projections ($r{=}8$, $\alpha{=}16$, dropout 0.1).
Training follows a two-stage schedule: epoch 1 optimizes regression heads only, while subsequent epochs jointly update heads and LoRA parameters.
Hyperparameters were selected via manual tuning on the development set. Best checkpoint was chosen by maximizing the sum of LQ and EXP Pearson correlations.

\section{Experiments}
\subsection{Experimental Setup}

Experiments are conducted on the SI corpus described in Section~3 using talk-level splits to ensure generalization to unseen talks. Training was performed on Google Colab using an NVIDIA L4 GPU for approximately 10 epochs, with each run completing within several hours depending on batch size and sequence length. 

Evaluation metrics include Pearson correlation ($r$) as the primary measure of ranking alignment with human ratings, Spearman correlation ($\rho$) for robustness to scale differences, prediction standard deviation to detect collapse, and cross-dimension correlation. Mean squared error (MSE) is additionally reported to capture absolute prediction deviation. Statistical significance is assessed using Fisher's $r$-to-$z$ transformation and bootstrap resampling (10{,}000 samples).

\subsection{Baselines}
Baselines are designed to isolate supervision structure under comparable backbone capacity: (i) \textbf{Frozen COMET-KIWI}: the pretrained \texttt{wmt22-cometkiwi-da} model applied without fine-tuning, with scalar predictions correlated separately with LQ and EXP; (ii) \textbf{Frozen Encoder + Linear Heads}: dual regression heads trained on frozen COMET-KIWI representations; (iii) \textbf{Single-Head Fine-Tuning}: LoRA-adapted scalar regression predicting the mean of LQ and EXP; (iv) \textbf{Prompt-Based LLM Evaluation}: structured zero-shot and few-shot rubric prompting with enforced scoring order (evaluated on the development set); (v) \textbf{Mean Baseline}: prediction of training-set mean scores.

\subsection{Human Reliability Analysis}
Given the subjective nature of SI evaluation, we quantify inter-rater reliability on overlapping dual-annotated segments.

\subsubsection{Inter-Rater Agreement}

\begin{table}[t]
 \caption{Human inter-rater reliability on dual-annotated segments.}
  \label{tab:human_ceiling}
  \centering
  \footnotesize
  \setlength{\tabcolsep}{2.0pt}
  \begin{tabular}{lccccc}
    \toprule
    Dim. & $r$ & 95\% CI & $\rho$ & ICC2 & ICC3 \\
    \midrule
    LQ  & .207 & [.107,.303] & .196 & .139 & .343 \\
    EXP & .274 & [.173,.369] & .291 & .140 & .430 \\
    \bottomrule
  \end{tabular}
\end{table}

On the dual-annotated subset ($n{=}367$ for LQ; $n{=}345$ for EXP), we estimate reliability using correlation- and variance-based measures. Because segments are annotated by different rater pairs, ICC is computed by treating the two ratings per segment as repeated measurements. ICC(2,1) captures absolute agreement, while ICC(3,1) captures consistency after rater scale shifts. Table~\ref{tab:human_ceiling} shows low absolute agreement but moderate consistency-level reliability.

\subsubsection{Inter-rater pairwise correlation}
To complement ICC, we compute pairwise Pearson correlations across rater pairs on overlapping subsets and average results across available combinations. Mean pairwise Pearson correlations are 0.264 (LQ), 0.286 (EXP), and 0.223 (LAT), providing a human--human reference ceiling for model correlation under subjective segment-level scoring.

\section{Results}
\subsection{Prompt-Based LLM Evaluation (Dev Set)}
\begin{table}[t]
  \caption{LLM evaluation behavior on the dev set.}
  \label{tab:llm_behavior}
  \centering
  \footnotesize
  \setlength{\tabcolsep}{2.5pt}
  \begin{tabular}{lcccc}
    \toprule
    Dim. & Human std & LLM std & Human $r$ & LLM $r$ \\
    \midrule
    LQ/EXP & .69/.74 & 1.07/1.27 & .56 & .90 \\
    \bottomrule
  \end{tabular}
\end{table}

Structured rubric prompting is evaluated on the development set ($n=87$). Despite explicit instructions enforcing dimension separation, prompt-based evaluation shows near-zero correlation with human ratings: zero-shot prompting yields negligible correlation, while few-shot prompting modestly improves LQ (Pearson $=0.13$) but not EXP (Pearson $=0.01$).

Table~\ref{tab:llm_behavior} highlights two failure modes: predictions exhibit substantially higher variance than human ratings and strong cross-dimensional coupling (corr $=0.90$ vs.\ human $=0.56$), indicating collapse toward a single latent quality signal despite structured prompting.

\subsection{Scalar Supervision Failure (Dev Set)}

We next examine scalar supervision by fine-tuning a single-head model to predict the mean of LQ and EXP. On the development set ($n=87$), correlations remain near zero (Pearson$(s,\mathrm{LQ})=0.092$, Pearson$(s,\mathrm{EXP})=-0.020$, Pearson$(s,\mathrm{combined})=0.050$), indicating that collapsing rubric dimensions into a single objective obscures ranking structure.

\subsection{Main results}

\begin{table}[t]
  \caption{Pearson correlation on development and test sets}
  \label{tab:main_results}
  \centering
  \begin{tabular}{lcc}
    \toprule
    Model & LQ ($r$) & EXP ($r$) \\
    \midrule
    \multicolumn{3}{c}{\textit{Development Set}} \\
    Prompt (zero-shot) & 0.054 & 0.041 \\
    Prompt (few-shot) & 0.130 & 0.010 \\
    Scalar fine-tune   & 0.092 & -0.020 \\
    \midrule
    \multicolumn{3}{c}{\textit{Test Set}} \\
    Mean baseline & 0.000 & 0.000 \\
    Frozen COMET-KIWI & 0.219 & 0.175 \\
    Dual-head (proposed) & \textbf{0.388} & \textbf{0.301} \\
    \bottomrule
  \end{tabular}
\end{table}

Table~\ref{tab:main_results} reports Pearson correlation with human ratings on development and test sets ($n{=}169$ for test). On the development set, prompt-based evaluation and scalar supervision yield near-zero correlations (LQ $r{\leq}0.130$, EXP $r{\leq}0.041$), indicating that both instruction-level prompting and scalar regression objectives fail to preserve rubric dimensionality.

On the held-out test set, the mean baseline shows no ranking ability, while frozen COMET-KIWI demonstrates moderate transfer (LQ $r{=}0.219$, EXP $r{=}0.175$), suggesting that translation-oriented representations partially capture SI quality signals. The proposed dual-head model substantially improves alignment with human ratings, achieving $r{=}0.388$ for LQ and $r{=}0.301$ for EXP, with gains over the frozen baseline statistically significant under bootstrap resampling ($p{<}0.05$).

Considering moderate consistency-level human reliability (ICC(3,1)=0.343 for LQ and 0.430 for EXP; Table~\ref{tab:human_ceiling}), these results indicate that structured supervision enables recovery of stable ranking signals under subjective segment-level annotation, positioning model performance within the human–human correlation range observed under subjective segment-level scoring.

\subsection{Qualitative Error Analysis}
\begin{table}[t]

\caption{Representative qualitative examples.}
\label{tab:qual_examples}
\centering
\small
\begin{tabular}{p{0.06\linewidth}p{0.30\linewidth}p{0.18\linewidth}p{0.15\linewidth}}
\toprule
ID & Interpretation (excerpt) & Gold (LQ/EXP) & Pred (LQ/EXP) \\
\midrule
E1 & ``...one becomes writer, one becomes prisoner.'' & 3 / 2 & 3 / 2 \\
E2 & ``Maybe it's about economic shift...'' & 2 / 2 & 2 / 2 \\
F1 & ``...use SDK to build process app...'' & 1 / 1 & 3 / 2 \\
F2 & ``...choose create new project...'' & 1 / 1 & 2 / 2 \\
\bottomrule
\end{tabular}
\end{table}

Representative examples (Table~\ref{tab:qual_examples}) illustrate correct separation of meaning transfer and delivery quality. Failure cases occur primarily in procedural segments containing multi-step instructions (F1–F2). In these examples, the interpretation preserves the overall workflow and action type but omits or distorts several individual steps, leading the model to assign high LQ despite incomplete content coverage. This pattern suggests that the model captures global procedural semantics while remaining insensitive to step-level information completeness.

\subsection{Dimension Disentanglement Analysis}

We examine cross-dimensional coupling on the test set, where the proposed model yields
$\mathrm{corr}(\hat{s}_{LQ}, \hat{s}_{EXP}) = 0.529$.
For comparison, human ratings exhibit
$\mathrm{corr}(s_{LQ}, s_{EXP}) = 0.56$,
while prompt-based LLM evaluation shows near-complete coupling
($\approx 0.90$, Table~\ref{tab:llm_behavior}). The proposed model therefore reproduces a coupling structure closely aligned with human judgment, in contrast to prompt-based evaluation that collapses both dimensions into a single latent quality signal. This finding indicates that parameter-level disentanglement preserves human-like partial dependence between meaning transfer and delivery quality while avoiding artificial over-coupling.

\section{Discussion}
Results indicate that the main bottleneck in automatic SI evaluation lies in supervision structure: prompt-based and scalar objectives collapse rubric dimensions, while multi-head supervision preserves human-like coupling. The weak association between perceived latency and objective delay suggests that temporal judgments reflect discourse-level synchrony, motivating future multimodal modeling. Limitations include the text-only setting, moderate dataset size, rater scale variability, and Chinese--English focus. We target stable ranking signals for formative assessment rather than absolute agreement. Labels, guidelines, splits, and code will be released where permitted; restricted materials remain subject to original licenses and consent constraints.

\section{Conclusion}

We presented a rubric-aligned framework for segment-level SI evaluation under professional analytic scoring. Using 1,101 annotated segments, we showed that prompt-based and scalar supervision collapse rubric dimensions, yielding weak alignment with human ratings. A dual-head COMET-KIWI model with LoRA improved correlation (Pearson $=0.388$ LQ, $0.301$ EXP) while preserving human-like cross-dimensional coupling, indicating recovery of stable ranking signals under subjective annotation. The approach enables scalable formative feedback and rubric-aligned benchmarking for SI training. Future work will extend to multimodal latency modeling and broader cross-lingual settings.

\section{Acknowledgments}

The authors thank Jason Zhang, a PhD student in Interpreting Studies at CUHK-Shenzhen, for valuable discussions on simultaneous interpreting annotation and evaluation design. We also thank Prof. Li Lan (School of Humanities and Social Science, CUHK-Shenzhen) for insightful feedback on rubric development and interpreting pedagogy perspectives. 

We are grateful to the interpreters and annotators who contributed their time and expertise to data collection and rating. We additionally acknowledge the providers of publicly available datasets and licensed conference materials used in this study.

This work was supported by Project W2531054 of the National Natural Science Foundation of China, and the Program for Guangdong Introducing Innovative and Entrepreneurial Teams.

\section{Generative AI Use Disclosure}

Generative AI tools were used for limited auxiliary purposes, including language editing during manuscript preparation, automated assistance in dataset organization, and code debugging support (e.g., via Cursor). All research design, experimental procedures, analyses, and conclusions were performed and verified by the authors, who take full responsibility for the content of this paper in accordance with ISCA policy.
\bibliographystyle{IEEEtran}
\bibliography{draft}

\end{document}